\documentclass[11pt]{article}
\usepackage{spconf,amsmath,graphicx}
\usepackage{float}
\usepackage{amssymb}
\usepackage{xcolor}
\usepackage{booktabs}
\usepackage{multirow}
\usepackage{enumitem}
\usepackage{mwe}
\usepackage{hyperref} 
\usepackage{tikz}
\usepackage{subcaption}
\usetikzlibrary{arrows.meta, positioning, shapes.geometric, calc}
\usepackage[numbers,sort&compress]{natbib}

\usepackage[letterpaper,
  left=0.63in,right=0.63in,
  top=1in,bottom=1.05in,
  columnsep=0.25in]{geometry}

\setlist{nosep, leftmargin=14pt}

\title{Integrating Neural Differential Forecasting with Safe Reinforcement Learning for Blood Glucose Regulation}

\name{Yushen Liu, Yanfu Zhang, Xugui Zhou}
\address{University of Virginia, College of William \& Mary, Louisiana State University}

\begin{document}
\maketitle

\begin{abstract}
Automated insulin delivery for Type~1 Diabetes must balance glucose control and safety under uncertain meals and physiological variability. While reinforcement learning (RL) enables adaptive personalization, existing approaches struggle to simultaneously guarantee safety, leaving a gap in achieving both personalized and risk-aware glucose control, such as overdosing before meals or stacking corrections. To bridge this gap, we propose TSODE, a safety-aware controller that integrates Thompson Sampling RL with a Neural Ordinary Differential Equation (NeuralODE) forecaster to address this challenge. Specifically, the NeuralODE predicts short-term glucose trajectories conditioned on proposed insulin doses, while a conformal calibration layer quantifies predictive uncertainty to reject or scale risky actions. In the FDA-approved UVa/Padova simulator (adult cohort), TSODE achieved 87.9\% time-in-range with less than 10\% time below 70\,mg/dL, outperforming relevant baselines. These results demonstrate that integrating adaptive RL with calibrated NeuralODE forecasting enables interpretable, safe, and robust glucose regulation.
\end{abstract}

\section{Introduction}

Type~1 Diabetes (T1D) is a chronic autoimmune disease affecting over 9.5 million people worldwide, characterized by the destruction of pancreatic $\beta$-cells and resulting in absolute insulin deficiency~\cite{alberti1998definition,davis2015cpeptide}. Individuals with T1D must carefully regulate blood glucose (BG) within a safe range (usually set between 70--180 mg/dL)~\cite{ada2022standards,zhou2021data,zhou2023hybrid}. Failure to do so can lead to acute hypoglycemia or hyperglycemia and long-term vascular and neurological complications, or even coma and death~\cite{mol2004hypoglycaemia,mynatt2010ubiquitous,zhou2022design}. Despite progress in continuous glucose monitoring (CGM) and insulin pumps, fully autonomous insulin delivery remains challenging due to delayed glucose–insulin dynamics, inter-patient variability, and uncertainty in meal timing and size~\cite{Jacobs2025AIDChallenges}.

Unsafe insulin dosing is a central barrier to reliable automation~\cite{Boughton2024AIDReview}. Small errors in timing or amount can trigger severe hypoglycemia, especially when controllers rely on imperfect glucose feedback or unannounced meals. Most reinforcement learning (RL), adaptive control, and MPC methods optimize overall glucose regulation but lack explicit safety mechanisms to constrain high-risk insulin actions~\cite{lee2023safe,zhu2024adaptive}. They also often ignore structured meal patterns, treating carbohydrate intake as random noise rather than a predictable disturbance that can be estimated or scheduled~\cite{jiang2022reinforcement}. As a result, these systems tend to react too late to postprandial glucose excursions or deliver excessive corrective doses.

To address these challenges, we develop a safety-aware RL controller that learns adaptive insulin dosing while certifying each proposed action against a predictive glucose model. A Neural Ordinary Differential Equation (NeuralODE) forecaster captures continuous-time glucose dynamics conditioned on insulin and meal context, enabling short-term (30-minute) trajectory prediction. A conformal calibration-based safety gate quantifies prediction uncertainty and scales or rejects insulin doses that could lead to hypoglycemia. Together, these components allow proactive and individualized control that remains safe under uncertain meal conditions.

Our framework, termed TSODE, integrates three synergistic components, including a Thompson Sampling policy for adaptive insulin selection, a NeuralODE forecaster for glucose trajectory prediction, and a conformal safety layer that enforces certified action filtering (see Fig. \ref{fig:pipeline} on the later page). Evaluated in the FDA-approved UVa/Padova Type 1 Diabetes Simulator~\cite{dallaman2014uva}, TSODE achieves higher time-in-range while reducing hypoglycemia risk more than 60\% compared to RL-control and model-predictive baselines.

\section{Related Work}
Classical glucose control algorithms such as proportional integral derivative (PID) and model predictive control (MPC) have demonstrated strong performance in maintaining near-normoglycemia~\cite{bequette2013challenges,kovatchev2009artificial,bagheri2022towards}. However, these approaches depend on explicit physiological models and manual parameter tuning, limiting their adaptability to patient variability and unannounced meals. Data-driven glucose forecasters based on deep learning~\cite{mirshekarian2022neural,smith2023short} can improve short-term prediction accuracy but cannot directly optimize sequential dosing decisions.

Offline reinforcement learning (RL) has recently emerged as a safer alternative for healthcare control problems where online exploration is risky~\cite{levine2020offline}. In the context of blood glucose regulation, several studies have shown that RL can enhance time-in-range (TIR) and personalization using only retrospective or simulated data. Examples include offline RL for safe insulin control~\cite{emerson2022offline}, distributional and uncertainty-aware RL for titration~\cite{desman2025glucose}, and deep actor–critic formulations for hybrid closed-loop systems~\cite{lee2023safe,zhu2024adaptive}.

While these approaches improve glycemic regulation, they typically focus on maximizing average performance without explicitly constraining unsafe actions such as overdosing or rapid correction stacking. Recent work \cite{zhou2025knowsafe,symeonidis2025deep} addresses this limitation but relies on multiple DNN models, increasing system complexity. The proposed TSODE framework integrates predictive safety certification via a single NeuralODE-based probabilistic forecaster and conformal calibration directly into a Thompson Sampling RL controller.

\section{Methodology}

\subsection{Overview}

\begin{figure}[!t]
    \centering
    \includegraphics[width=1\linewidth]{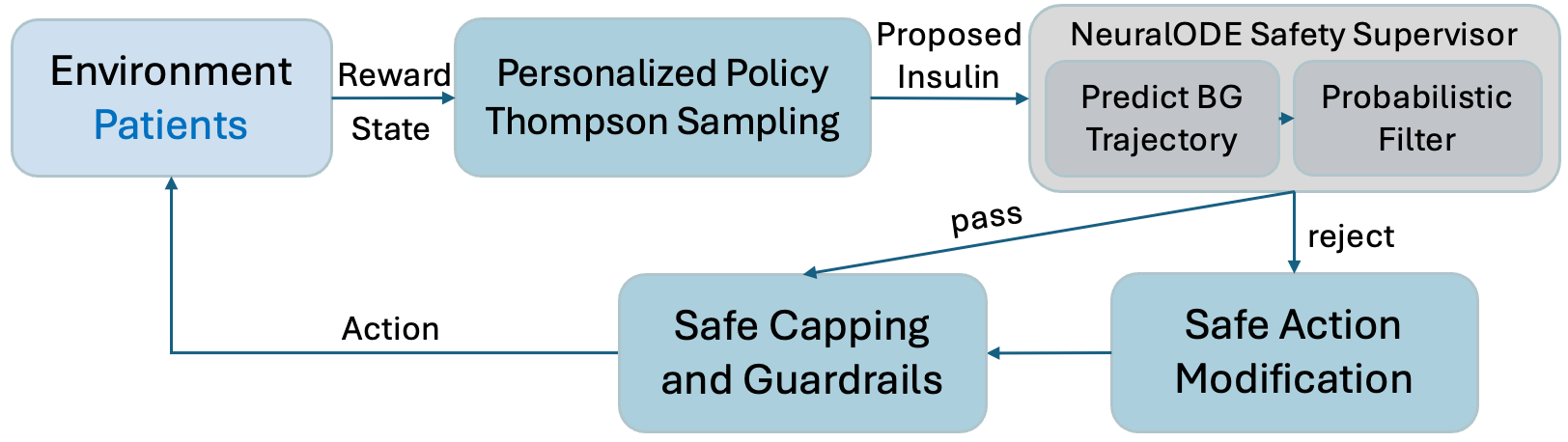}
    \caption{The model architecture TSODE}
    \label{fig:pipeline}
\end{figure}

We design a discrete-action closed-loop controller that combines Thompson Sampling reinforcement with NeuralODE forecasting-based safety supervision, as in Fig.~\ref{fig:pipeline}. The controller operates in discrete time. At each decision step, it observes recent glucose and contextual features, proposes an insulin action through a Thompson Sampling policy, and evaluates its safety using a probabilistic forecaster based on neural ordinary differential equations. Unsafe proposals are adaptively corrected through a constrained optimization step, while additional rule-based guardrails enforce hard physiological limits.

\subsection{State and Action Space}

At each decision step $t$, the controller forms a state vector $\mathbf{x}_{t-H+1:t}$ from recent glucose readings, insulin delivery (IOB), carbohydrate estimates (COB), time-of-day, and risk metrics.  
The action space $\mathcal{A}$ is a discrete grid of insulin bolus increments within a physiologically safe range.

To enhance meal responsiveness, a feedforward pre-bolus is added during scheduled or predicted carbohydrate intake, computed from carbohydrate sensitivity, and merged with the RL-proposed action. The combined insulin command is projected onto $\mathcal{A}$ before passing to the safety layer. This compact formulation preserves physiological interpretability while allowing flexible feature design, action resolution, and feedforward adaptation.

\subsection{Thompson Sampling Policy}

Glucose level $g_t$ and short-term trend $\dot{g}_t$ are discretized into bins to form a finite state $s=(b_{\text{BG}},b_{\text{TR}})$, where $b_{\text{BG}}$ indexes a BG bin in $[40,300]$\,mg/dL with 20\,mg/dL spacing and $b_{\text{TR}}$ indexes a trend bin over $\{-\infty,\allowbreak -2,\allowbreak -1,\allowbreak -0.3,\allowbreak 0.3,\allowbreak 1,\allowbreak 2,\allowbreak \infty\}$. 
The trend $\dot g_t$ is computed as the mean per-step change over the last three glucose samples.

For each state–action pair $(s,a)$, the controller maintains numerically stable online estimates of the mean and variance of the shaped reward.
Action selection follows the standard Thompson Sampling rule, where each action’s expected reward is modeled as Gaussian, a sample $\tilde{r}_a$ is drawn from the posterior of each action, and the controller selects
\begin{align}
a_t = \arg\max_a \tilde{r}_a\quad
\tilde{r}_a \sim \mathcal{N}(\mu_a, \sigma_a^2/n_a)
\label{eq:2}
\end{align}
where $n_a$ denotes the number of observations for action $a$. 
During evaluation, the policy acts greedily using $\arg\max_a \mu_a$. 
This stochastic formulation naturally balances exploration and exploitation without additional hyperparameters or explicit uncertainty calibration.

\subsection{Neural Ordinary Differential Equation Forecaster}

To model continuous-time glucose–insulin dynamics, we employ a NeuralODE forecaster that predicts short-term glucose trajectories conditioned on recent physiological context and proposed insulin actions. At each control step, the model receives a sliding history window of standardized features 
(BG, IOB, COB, time-of-day, and risk indices) and encodes it through a gated recurrent unit (GRU) to produce a latent initial state $\mathbf{z}_0$.

The latent state evolves under a NeuralODE $\dot{\mathbf{z}}=f_\theta(\mathbf{z})$, parameterized by a lightweight multilayer perceptron. Integration over the prediction horizon yields latent trajectories $\mathbf{z}_{1:K}$, which are decoded into glucose mean and variance estimates $(\hat{\mu},\hat{\sigma}^2)$. 

We train the model using a heteroscedastic Gaussian negative log-likelihood (NLL) loss
\begin{align}
\mathcal{L}_{\text{NLL}}
=\tfrac{1}{2}\big(\log\hat{\sigma}^{2}
+\tfrac{(y-\hat{\mu})^2}{\hat{\sigma}^{2}}\big)
\label{eq:1}
\end{align}
which allows the forecaster to learn data-dependent uncertainty. This adaptive weighting improves stability on noisy sensor data and accelerates convergence on traces with abrupt postprandial excursions or recovery slopes. Empirically, the NLL objective produces smoother yet more physiologically realistic trajectories than mean-squared error, capturing sharper glucose inflection points without overfitting transient fluctuations. The resulting NeuralODE forecaster therefore provides both predictive accuracy and calibrated uncertainty estimates, forming the probabilistic foundation for the conformal safety filter described next.

\subsection{Probabilistic Safety Filter}
Given a proposed bolus \(u\), the NeuralODE forecaster predicts a glucose trajectory \(\mu_{1:K}\), where \(\mu_k\) denotes the predicted blood glucose (mg/dL) at future step \(k\). Hypoglycemia risk \cite{zhou2021data,zhou2023hybrid} is summarized using a weighted trajectory average and a horizon-level slope:
\begin{align}
W(\mu) &= \sum_{k=1}^{K} w_k\,\mu_k, \qquad
S(\mu) = \frac{\mu_K-\mathrm{BG}_t}{K\,\Delta t}
\end{align}
where \(K\) is the prediction horizon, \(w_k\) are exponentially decaying weights emphasizing near-term predictions, \(\mathrm{BG}_t\) is the current glucose value, and \(\Delta t\) is the control interval.

Safety is enforced by requiring both statistics to exceed predefined thresholds with high probability:
\begin{align}
\Pr\!\big(W(\mu)\ge L,\; S(\mu)\ge -\gamma\big) \;\ge\; 1-\alpha,
\end{align}
These constraints enforce high-probability lower bounds on predicted glucose level and descent rate, respectively. Uncertainty is handled via conformal calibration using empirical residual quantiles. If tests fail, a 1-D bisection over \(u\in[0,u_{\mathrm{prop}}]\) returns the largest safe dose, else \(u=0\). When BG is high and rising, the forecast gate is bypassed. Final delivery passes deterministic guardrails: glucose/trend thresholds, IOB caps, and a late-night cap.

\begin{table*}[t]
\centering
\caption{Comparison of Algorithm Performance (Means over 14 Days)}
\label{tab:comparison}
\resizebox{\textwidth}{!}{%
\begin{tabular}{lcccccccccccc}
\toprule
\multirow{2}{*}{\textbf{Patient}} &
\multicolumn{3}{c}{\textbf{Meal-Bolus Baseline}} &
\multicolumn{3}{c}{\textbf{PID}} &
\multicolumn{3}{c}{\textbf{TSMPC}} &
\multicolumn{3}{c}{\textbf{TSODE (OUR METHOD)}} \\
\cmidrule(lr){2-4} \cmidrule(lr){5-7} \cmidrule(lr){8-10} \cmidrule(lr){11-13}
 & \textbf{TIR\%} & \textbf{Time$<$70\%} & \textbf{Mean BG (mg/dL)} 
 & \textbf{TIR\%} & \textbf{Time$<$70\%} & \textbf{Mean BG (mg/dL)}
 & \textbf{TIR\%} & \textbf{Time$<$70\%} & \textbf{Mean BG (mg/dL)}
 & \textbf{TIR\%} & \textbf{Time$<$70\%} & \textbf{Mean BG (mg/dL)} \\
\midrule
adult\#001 & 21.19 & 0.00 & 209.6 & 75.09 & 24.82 & 96.3 & 73.71 & 26.29 & 97.6 & \textbf{84.60} & 9.11 & 121.9 \\
adult\#002 &  1.55 & 0.00 & 206.9 & 86.64 & 13.36 & 101.1 & 78.96 & 21.04 & 101.6 & \textbf{87.80} & 7.60 & 120.9 \\
adult\#005 &  2.04 & 0.00 & 215.3 & 89.25 &  9.99 & 109.8 & 87.90 & 10.70 & 112.7 & \textbf{91.28} & 1.68 & 132.7 \\
\midrule
\textbf{Mean (adult\#001–010)}
& 15.04 & 0.00 & 211.7
& 77.58 & 21.76 & 98.5
& 75.89 & 23.10 & 100.5
& \textbf{83.75} & 8.83 & 123.8 \\

\bottomrule
\end{tabular}%
}
\end{table*}

\section{Experimental Evaluations}

\subsection{Experimental Setting}
\noindent\textbf{Data Description.} 
All experiments were performed using a closed-loop testbed \cite{zhou2022design} built on the UVa/Padova Type~1 Diabetes simulator. The simulator models glucose-insulin dynamics for virtual adults with individualized metabolic parameters.
Real patient experiments are not conducted due to the safety-critical nature of insulin dosing.

\noindent\textbf{Experiment Setup.}
Each virtual adult patient was simulated for 44 consecutive days in the simulation. The first 30 days were used for policy warm-up and the last 14 for closed-loop evaluation. Every $\Delta t=3$\,min, the controller observed a normalized feature vector $x_t$ over a sliding window of $H=10$ steps (30\,min). The discrete action set
$\mathcal{A}=\{0,0.2,\ldots,3.0\}$\,U
represents physiologically safe bolus increments (in international units), with a short refractory period preventing rapid re-dosing. All controllers followed identical meal schedules (08:00, 12:30, 16:00, 19:00; 50-70-15-60\,g) and were simulated with fixed random seeds for reproducibility. An open-source implementation is available at \href{https://github.com/Hudl1e/TSODE_control.git}{GitHub}.

\noindent\textbf{Benchmark Algorithms.}
We compared TSODE with three standard baselines:
(i) a rule-based Meal-Bolus controller using fixed insulin-to-carbohydrate ratio (ICR) and basal rate,
(ii) a classical PID loop maintaining BG near 120 mg/dL, and
(iii) a hybrid MPC-based controller with Thompson Sampling for adaptive action selection (TSMPC), serving as a predictive-control baseline with analytic constraints and no learned uncertainty modeling.

\subsection{Results}

We evaluate the proposed TSODE controller against multiple baseline algorithms to assess both glycemic performance and safety under realistic meal scenarios. Performance was quantified using standard clinical metrics: TIR, time below 70\,mg/dL, and mean BG.

\begin{figure}[!t]
    \centering
    \begin{subfigure}[b]{0.47\columnwidth}
        \centering
        \includegraphics[width=\linewidth]{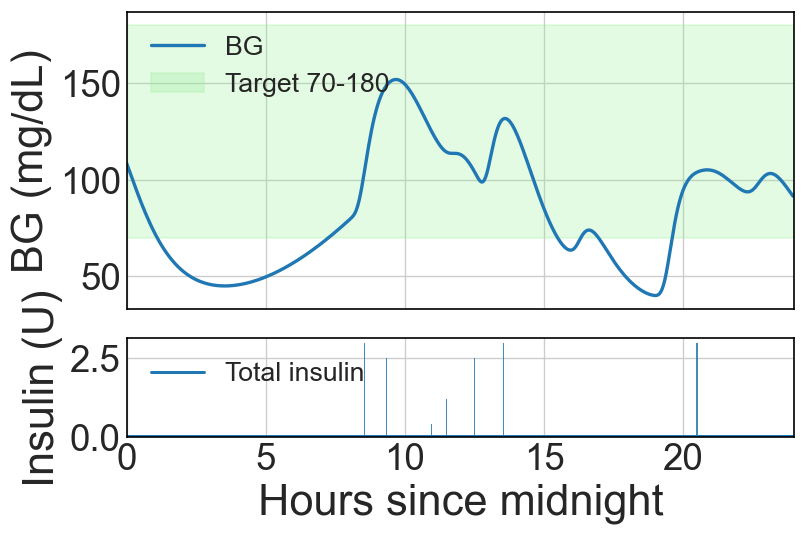}
        \label{fig:day-sim}
    \end{subfigure}
    \hfill
    \begin{subfigure}[b]{0.47\columnwidth}
        \centering
        \includegraphics[width=\linewidth]{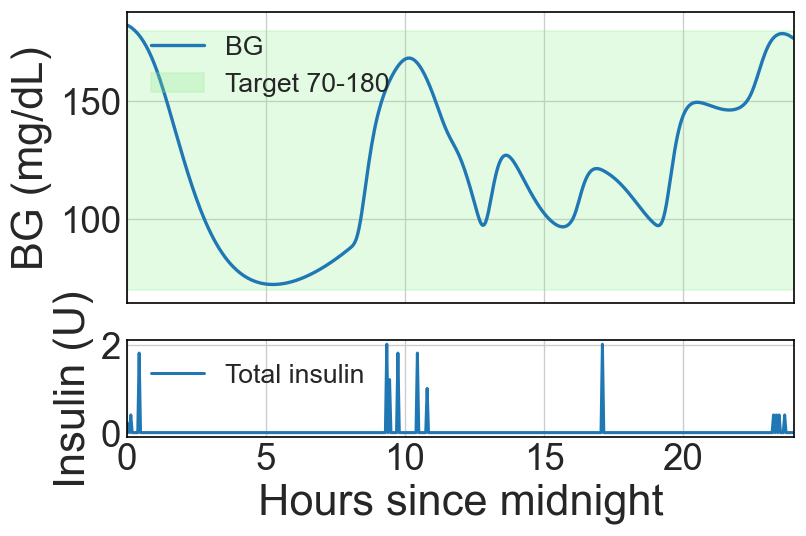}
        \label{fig:tir}
    \end{subfigure}

    \vspace{-10pt}
    \caption{Comparison of PID (left) and TSODE (right) over a 24-hour simulation for adult\#001.}
    \label{fig:results}
\end{figure}

\begin{figure}[!t]
    \centering
    \includegraphics[width=1\linewidth]{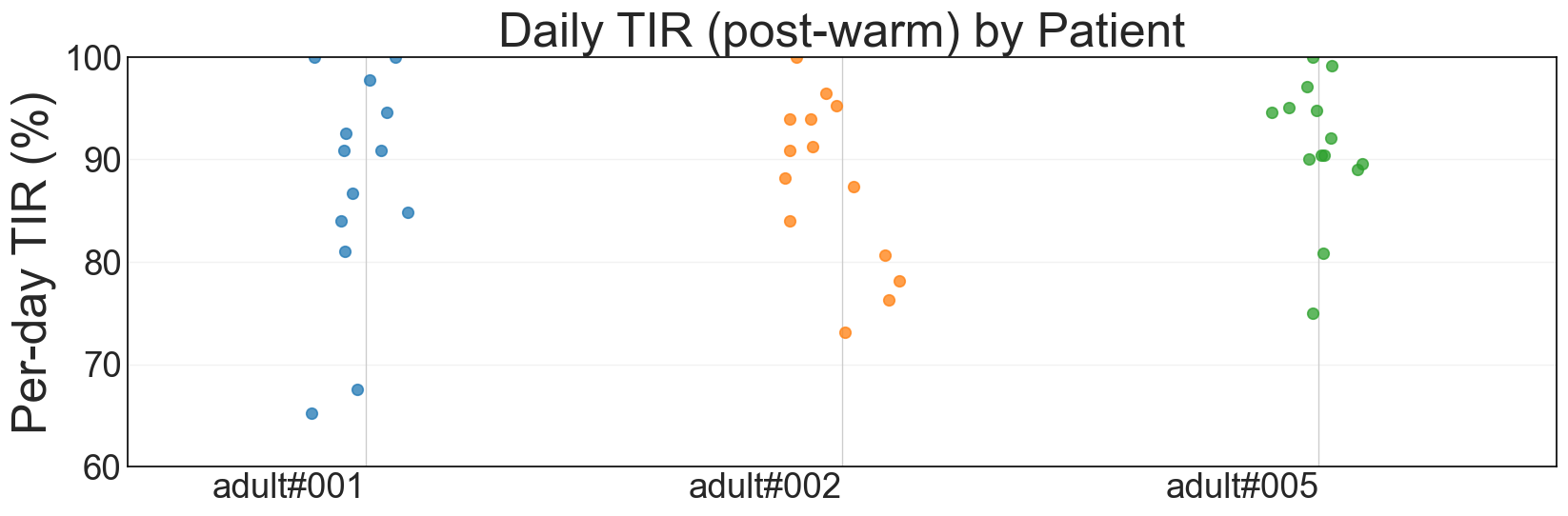}
    \caption{Daily TIR by Patient Using TSODE}
    \label{fig:result2}
\end{figure}

Figs.~\ref{fig:results} and~\ref{fig:result2} illustrate the closed-loop performance of the proposed TSODE controller. Fig.~\ref{fig:results} compares a representative 24-hour simulation against a conventional PID controller for the same virtual adult. 
While PID shows larger glucose excursions and periods of suboptimal regulation, TSODE maintains levels tightly within the target range of 70–180\,mg/dL, achieving full time-in-range with no hypoglycemia. 
Fig.~\ref{fig:result2} extends this comparison across three adult subjects, where daily TIR values consistently cluster between 85\% and 100\%, reflecting stable glycemic control and strong generalization despite variability in metabolism and insulin sensitivity. 
Together, these results demonstrate that TSODE achieves safe, adaptive regulation superior to conventional controls.

Table~\ref{tab:comparison} summarizes per-patient results and aggregate performance over 14 evaluation days. 
Across three representative adults and a six-subject cohort mean (adult\#001, \#002, \#005, \#008–\#010), TSODE consistently achieves the highest TIR while limiting time below 70\,mg/dL to under 10\%, substantially outperforming PID and TSMPC, which incur up to 26\% hypoglycemia time. 
These results demonstrate that TSODE maintains strong glycemic control and safety across both individual cases and heterogeneous patient dynamics. By combining meal-aware adaptation with NeuralODE forecasting, TSODE anticipates downward trends and caps unsafe doses, enhancing both performance and safety in autonomous insulin delivery.

\noindent\textbf{Model Transferability and Generalization.}
To evaluate cross-patient generalization, the TSODE controller was trained on adults\#001 and \#002 and tested on an unseen subject (adult\#005). As summarized in Table \ref{tab:comparison}, the baseline achieved only 21.2\% TIR, while the PID improved TIR to 75.1\% but incurred 24.8\% time below 70\,mg/dL. In contrast, TSODE maintained a high TIR of 88.9\% with only 3.8\% time $<$ 70\,mg/dL, achieving comparable control with substantially reduced hypoglycemia risk. These results highlight the controller’s ability to generalize across unseen patients while preserving safety through NeuralODE-based probabilistic supervision.

\section{Conclusion}
We presented TSODE, a safety-aware controller that integrates Thompson Sampling reinforcement learning with NeuralODE-based probabilistic forecasting for automated insulin delivery.
TSODE adaptively regulates glucose while certifying each insulin action through conformal safety checks and guardrail constraints.
In the FDA-approved UVa/Padova simulator, TSODE achieved a mean time-in-range of 83.75\% with only 8.83\% time below 70\,mg/dL, representing a 6--8\,pp improvement in TIR and a 60\% reduction in hypoglycemia compared to PID and TSMPC baselines.
These results demonstrate that probabilistic safety filtering substantially enhances both stability and safety of closed-loop glucose control.

\newpage
{\footnotesize
\bibliographystyle{IEEEtranN}
\bibliography{references}
}

\end{document}